\documentclass[a4paper,conference]{IEEEtran}
\usepackage{cite}
\usepackage{graphicx}
\usepackage{tabularx}
\usepackage{amsmath} 
\usepackage{epsfig} 
\usepackage{url}
\usepackage{makecell}
\usepackage{subfigure} 
\def\BibTeX{{\rm B\kern-.05em{\sc i\kern-.025em b}\kern-.08em
    T\kern-.1667em\lower.7ex\hbox{E}\kern-.125emX}}

\usepackage[table]{xcolor}
\usepackage{collcell}
\usepackage{hhline}
\usepackage{pgf}
\usepackage{multirow}
\usepackage[english]{babel}
\def\colorModel{hsb} %You can use rgb or hsb

\newcommand\ColCell[1]{
  \pgfmathparse{#1<50?1:0}  %Threshold for changing the font color into the cells
    \ifnum\pgfmathresult=0\relax\color{white}\fi
  \pgfmathsetmacro\compA{0}      %Component R or H
  \pgfmathsetmacro\compB{#1/100} %Component G or S
  \pgfmathsetmacro\compC{1}      %Component B or B
  \edef\x{\noexpand\centering\noexpand\cellcolor[\colorModel]{\compA,\compB,\compC}}\x #1
  } 
\newcolumntype{E}{>{\collectcell\ColCell}m{1.2cm}<{\endcollectcell}}  %Cell width

\usepackage[colorlinks]{hyperref}

\begin{document}
\title{RWF-2000: An Open Large Scale Video Database for Violence Detection}
\author{\IEEEauthorblockN{Ming Cheng}
\IEEEauthorblockA{Duke Kunshan University\\
Kunshan, China\\
ming.cheng@dukekunshan.edu.cn}
\and
\IEEEauthorblockN{Kunjing Cai}
\IEEEauthorblockA{Sun Yat-sen Universit\\
Guangzhou, China\\
caikj3@mail2.sysu.edu.cn}
\and
\IEEEauthorblockN{Ming Li}
\IEEEauthorblockA{Duke Kunshan University\\
Kunshan, China\\
ming.li369@dukekunshan.edu.cn}
}

\maketitle

\begin{abstract}
In recent years, surveillance cameras are widely deployed in public places, and the general crime rate has been reduced significantly due to these ubiquitous devices. Usually, these cameras provide cues and evidence after crimes are conducted, while they are rarely used to prevent or stop criminal activities in time. It is both time and labor consuming to manually monitor a large amount of video data from surveillance cameras. Therefore, automatically recognizing violent behaviors from video signals becomes essential. This paper summarizes several existing video datasets for violence detection and proposes the RWF-2000 database with 2,000 videos captured by surveillance cameras in real-world scenes. Also, we present a new method that utilizes both the merits of 3D-CNNs and optical flow, namely Flow Gated Network. The proposed approach obtains an accuracy of 87.25\% on the test set of our proposed database. The database and source codes are currently open to access \footnote{\url{https://github.com/mchengny/RWF2000-Video-Database-for-Violence-Detection}}.
\end{abstract}

\IEEEpeerreviewmaketitle

\section{Introduction}
Recently, video-based violent behavior detection has attracted more and more attention. There is an increasing number of surveillance cameras in public places, collecting evidence and deter potential criminals. However, it is too expensive to monitor a large amount of video data in real-time manually. Thus, automatically recognizing criminal scenes from videos becomes essential and challenging.

Generally, the definition of video-based violence detection is detecting violent behaviors in video data. It is a subset of human action recognition that aims at recognizing common human actions. Compared to still images, video data has additional temporal sequences. A set of consecutive frames represent a continuous motion, and neighboring frames contain redundant information due to high inter-frame correlation. Thus, many researchers devote to fuse both spatial and temporal information properly.

Some earlier methods rely on detecting the presence of highly relevant objects (e.g., gun shooting, blaze, blood, blast) rather than directly recognizing the violent events \cite{r5, r6, r7}. In 2011, Nievas et al. firstly released two video datasets for violence detection, namely the Hockey Fight dataset \cite{nievas2011hockey} and the Movies Fight dataset \cite{nievas2011movies}. Later, Hassner et al. \cite{r3} proposed the Crowd Violence dataset in 2012. Since then, most works turn to develop methods that directly recognize violence in videos.

\begin{figure}[t]
  \includegraphics[width=\linewidth]{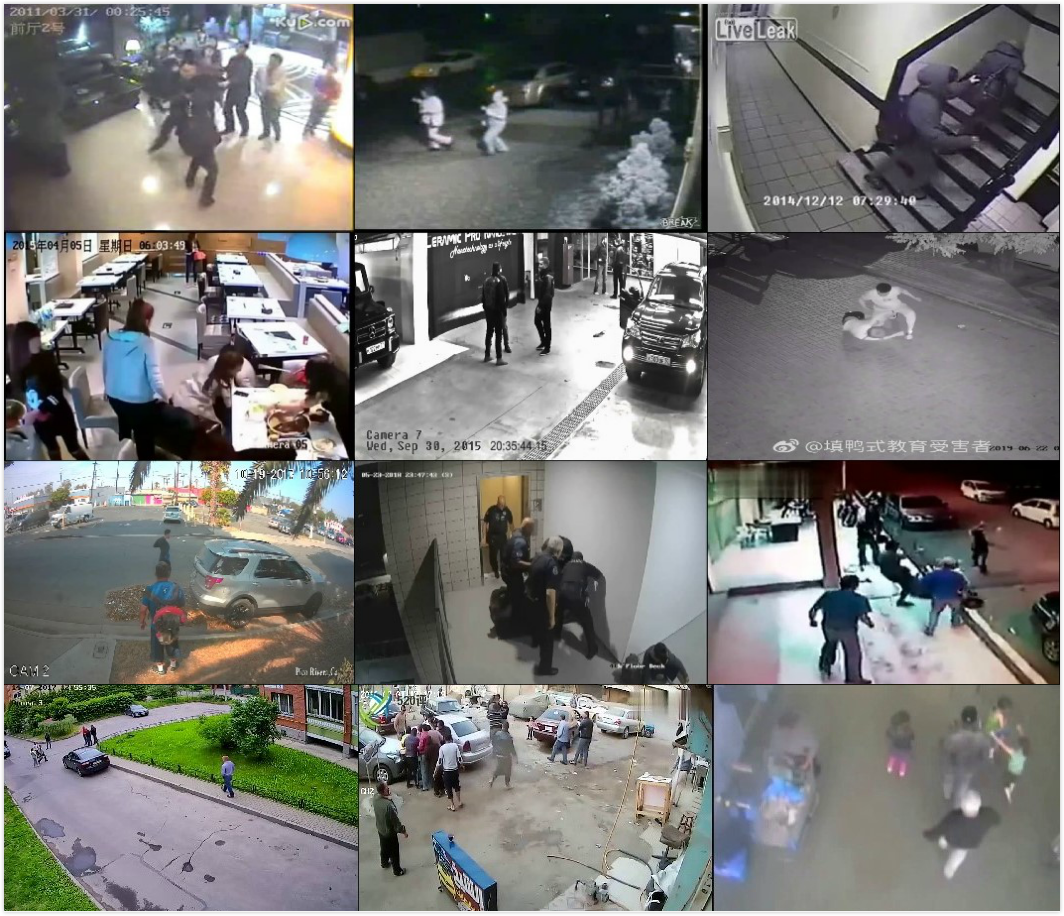}
  \caption{Gallery of the RWF-2000 database.}
  \label{gallery}
\end{figure}

There are mainly two categories of methods to recognize human action in videos: the traditional feature extraction with shallow classifiers and the end-to-end deep learning framework. In traditional methods, researchers mainly devote to build powerful video descriptors by feature extraction algorithms and feed them to a classifier (e.g., SVM). Based on this principle, many classical methods are presented: ViF \cite{r3}, STIPs \cite{r8}, SCOF \cite{r10}, iDT \cite{r11}, etc. While in recent years, more and more end-to-end deep learning based methods are proposed, e.g., two-stream method \cite{r14}, ConvLSTM \cite{r13}, C3D \cite{r1}, TSN \cite{r4}, ECO \cite{r2}. Currently, most state-of-the-art results are obtained through deep learning based methods.

\begin{table*}[t]
  \centering
  \renewcommand\arraystretch{1.25}
  \caption{Comparisons between the RWF-2000 and the previous datasets. The 'Natural' represents that videos are from realistic scenes, but recorded by hybrid types of devices (e.g., mobile cameras, car-mounted cameras). }
  \label{previous_datasets}
  \resizebox{\textwidth}{!}{
  \begin{tabular}{|c|c|c|c|c|c|c|}
    \hline
    \textbf{Authors}  & \textbf{Dataset} & \textbf{Data Scale} & \textbf{Length/Clip (sec)} & \textbf{Resolution} & \textbf{Annotation} & \textbf{Scenario} \\
    \hline
    Blunsden et al. \cite{behave} & BEHAVE & 4 Videos (171 Clips) & 0.24-61.92 & 640$\times$480 & Frame-Level & Acted Fights \\
    Rota et al. \cite{redid} & RE-DID & 30 Videos & 20-240 &  1280$\times$720 & Frame-Level & Natural \\
    Demarty et al.\cite{vsd} & VSD & 18 Movies (1,317 Clips) & 55.3-829.4 & Variable & Frame-Level & Movie \\
    Perez et al. \cite{cctv_fight} & CCTV-Fights & 1,000 clips & 5-720 & Variable & Frame-Level & Natural \\
    Nievas et al. \cite{nievas2011hockey} & Hockey Fight & 1,000 Clips & 1.6-1.96 & 360$\times$288 & Video-Level & Hockey Games \\
    Nievas et al. \cite{nievas2011movies} & Movies Fight & 200 Clips & 1.6-2 & 720$\times$480 & Video-Level & Movie \\
    Hassner et al. \cite{r3} & Crowd Violence & 246 Clips & 1.04-6.52 & Variable & Video-Level & Natural \\
    Yun et al. \cite{sbu} & SBU Kinect Interaction & 264 Clips & 0.67-3 & 640$\times$480& Video-Level & Acted Fights \\ 
    Sultani et al. \cite{ucf-crime} & UCF-Crime & 1,900 Clips & 60-600 & Variable & Video-Level & Surveillance \\
    Ours & RWF-2000 & 2,000 Clips & 5 & Variable & Video-Level & Surveillance \\
  \hline
  \end{tabular}}
\end{table*}

Although some video datasets for violence detection already exist, they still have drawbacks of small scale, reduced diversity, and low image resolution. Moreover, some related datasets with high image quality come from movies, which are not close enough to real-world scenes. To solve insufficient high-quality data from real violent activities, we collect a new video dataset (RWF-2000) and freely release it to the research community. This dataset has a large scale with 2,000 clips extracted from surveillance videos. Besides, we present a novel model with a self-learned pooling mechanism, which could adopt both appearance features and temporal features well.

\section{Related Work}
\subsection{Previous Datasets}
\label{other_datasets}

According to the annotation method, there are mainly two kinds of video datasets for violence detection: trimmed and untrimmed. The videos in trimmed datasets are all short clips with a length of several seconds, and each one of them has a video-level annotation. While the videos in untrimmed datasets usually have a longer duration. Furthermore, the start time and end time of violent activities have frame-level annotations. Table \ref{previous_datasets} shows the comparison between our proposed RWF-2000 dataset and previous others. 

Blunsden et al. \cite{behave} propose the BEHAVE dataset for multi-person behavior classification. This dataset consists of 2 to 5 people in one or two interacting groups. Invited actors perform ten group behaviors (InGroup, Approach, WalkTogether, Meet, Split, Ignore, Chase, Fight, RunTogether, Following). There are 171 interacting clips recorded by four surveillance videos from different shooting angles.

Nievas et al. present two video datasets for violence detection, namely the Movies Fight \cite{nievas2011movies} and the Hockey Fight \cite{nievas2011hockey}. The Movies Fight dataset has 200 clips extracted from short movies. The number of videos is insufficient nowadays. In the Hockey Fight dataset, there are 1,000 clips captured in hockey games of the National Hockey League. One of its disadvantages is the lack of diversity because all the videos are captured in a single scene. Both of these two datasets have video-level annotations.

Hassner et al. \cite{r3} collect 246 videos with or without violent behaviors, namely the Crow Violence dataset. This dataset aims at recognizing violence in crowded scenes. The length of videos is between 1.04 seconds and 6.53 seconds. The characteristic of this dataset is its overcrowded scenes but low image quality.

Compared to recognize violence by only RGB data, Yun et al. \cite{sbu} present the first violence dataset in the form of RGB-D data. There are eight interactions performed by actors (approaching, departing, pushing, kicking, punching, exchanging objects, hugging, and shaking hands). Using the Microsoft Kinect sensor, 21 pairs of two-actor interactions performed by seven participants are recorded. Also, this dataset provides video-level annotations and gives the joints data of the human skeleton.

Rota et al. \cite{redid} present the RE-DID (Real-Life Events-Dyadic Interactions) dataset. This dataset is composed of 30 videos with high-quality resolution up to 1280$\times$720.  Videos captured by car-mounted cameras account for 25\% of the total. Other types of devices take the rest (e.g., mobile phone). Moreover, the bounding boxes of participants with relative IDs, the position of the interpersonal spaces are both provided, which is useful for in-depth analysis.

Demarty et al. \cite{vsd} establish the VSD dataset. The dataset consists of 1,317 clips from 18 selected Hollywood movies (e.g., Kill Bill, Fight Club). The definition of violent content in the film involves blood, fights, fire, guns, cold weapons, car chases, gory scenes, gunshots, explosions, and screams. Each frame in this dataset is annotated as violence or non-violence by these criteria.

Those above datasets are mainly composed of videos captured in a single scene, performed by actors, or extracted from edited movies. Only a few parts of them are from real events. To make the trained model more practical, Sultani et al. \cite{ucf-crime} present the UCF-Crime dataset, which contains 1,900 videos recorded by real-world surveillance cameras. This video dataset is preliminarily designed to detect 13 real anomalies, including abuse, arrest, arson, assault, accident, burglary, explosion, fighting, robbery, shooting, stealing, shoplifting, and vandalism. While there are still some limits, videos in this dataset usually have a long duration ranging from 1 to 10 minutes, but they have only video-level annotations. Hence, detecting violent activities from a long video is very tough.

CCTV-Fights \cite{cctv_fight} is the latest large-scale video dataset for violence detection. It consists of 1,000 videos containing violent activities, and each violent activity is annotated with the start and end time. Only 280 videos are from real-world surveillance cameras, and 720 are taken by other types of devices, including mobile cameras, car cameras, drones, or helicopters. Only a small part of these videos comes from real surveillance cameras.

Summarizing these proposed datasets, each of them has at least one or more of the following limits:
\begin{itemize}
  \item low image quality;
  \item lack of sufficient data amount;
  \item videos with long duration but crude annotations;
  \item hybrid sources of videos that are not close enough to realistic violence.
\end{itemize}

To cope with the above issues, we collect a new RWF-2000 (Real-World Fighting) dataset from the YouTube website, consisting of 2,000 trimmed video clips captured by surveillance cameras from real-world scenes. Figure \ref{gallery} demonstrates some video examples in the RWF-2000 dataset. The detailed description of our proposed dataset will be introduced in Section \ref{data_collection}.  

\subsection{Previous Methods}
Traditional methods usually try to find a powfer feature extraction algorithm and implement a machine learning based classifier to complete the violence detection task. Some classical models involve space-time interest points (STIP) \cite{r8}, Harris corner detector \cite{d1}, improved dense trajectory (iDT) \cite{r11}, motion scale-invariant feature transform (MoSIFT) \cite{MoSIFT}, etc.

Chen et al. \cite{r6} explore the way to detect violence from compressed video data. They utilize the encoded motion vector from the MPEG-1 video data as extracted features, representing the data component with high motion intensity. By this data format, they build a classifier to detect the presence of blood as violence.

Hassner et al. \cite{r3} present the violent flows (ViF) descriptors for violence detection. This method can utilize the magnitude series of optical flow over time to detect violent activities. Later, Gao et al. \cite{d2} improve this method by introducing the orientation of the violent flow features into the ViF descriptor, namely the oriented violent flows (OViF).

Not limited to the field of violence recognition, the improved dense trajectory (iDT) \cite{r11} features has made a significant performance in generic human action recognition. Bilinski et al. \cite{d3} introduce iDT features into the violence recognition. Also, they present an improved Fisher encoding method that is powerful to encode the spatial-temporal information in a video.

With the fast development of deep learning in recent years, methods based on convolutional neural networks have become the mainstream in video-based action recognition (e.g., TSN \cite{r4}, ECO \cite{r2}, C3D \cite{r1}). Many researchers attempt to address the violence recognition task by introducing deep learning based methods \cite{d4, zhou2017violent, d5,d6, convlstm-violence}.

Dong et al. \cite{d6} present a multi-stream deep neural network for inter-person violence detection from videos. The model adopts raw videos, optical flows, and acceleration flow maps as three input branches. Followed by the Long Short Term Memory (LSTM) network, a score-level fusion is employed to adopt the three streams to predict a final result for violence detection.

\begin{figure}[t]
  \centering
  \includegraphics[width=\linewidth]{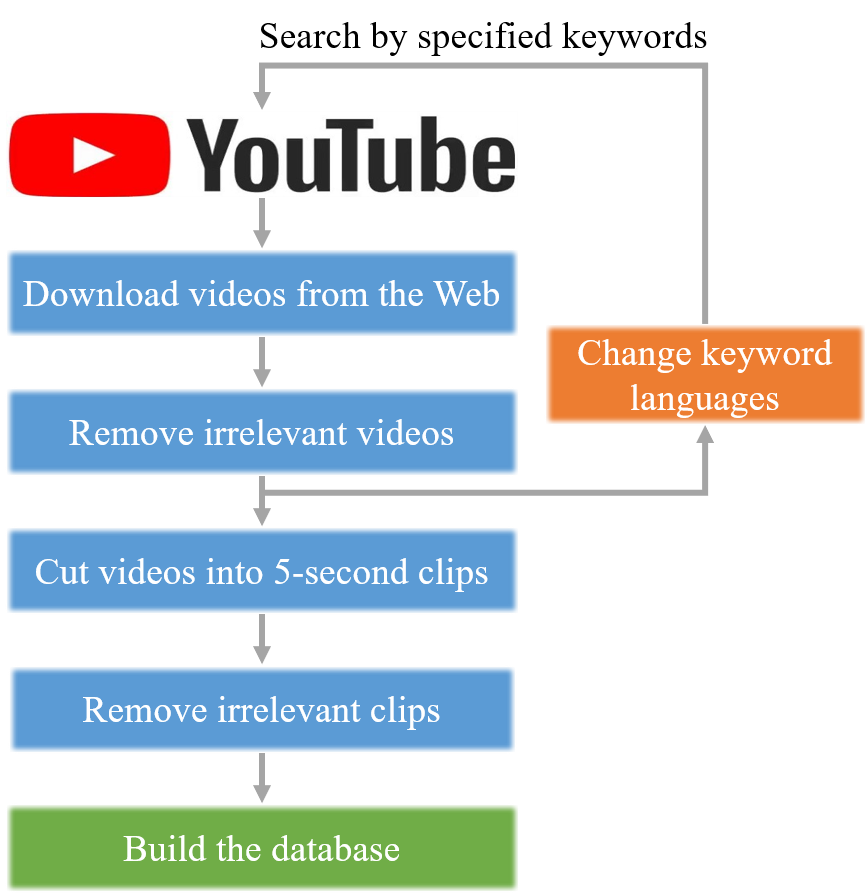}
  \caption{Pipeline of data collection.}
  \label{pipline}
\end{figure}

Sudhakaran et al. \cite{convlstm-violence} combine the advantages of both convolutional neural networks and recurrent neural networks to do violence detection. The ConvLSTM network \cite{d8} is modified to take the difference of adjacent frames as input instead of to give a sequence of raw frames. The convolutional neural network is deployed to extract spatial features at the frame-level, and the following LSTM network is used to model the temporal relationships.  

Zhou et al. \cite{zhou2017violent} build a FightNet to discover the violent activities in videos by multimodal data (e.g., RGB images, optical flow images, and acceleration images). They firstly pre-train the proposed model on a generic action recognition dataset named UCF101 \cite{d7} and then fine-tune it on the Crowd Violence dataset \cite{r3}. At the same time, it is found that the model is easy to fall into an over-fitting problem because of high model complexity under insufficient violence data.

In summary, deep learning based approaches usually outperform traditional feature extraction based models. Besides, most state-of-the-art results utilize multi-channel inputs (e.g., raw RGB images, optical flows, acceleration maps). At the same time, complex models are not very robust against Over-Fitting. In this paper, we only adopt RGB images and optical flows to build the neural network, which can process the spatial and temporal information. Furthermore, our proposed Flow-Gated architecture can reduce temporal channels of input videos by self-learning, instead of traditional pooling strategies. More details of the architecture will be described in Section \ref{model_structure}.

\section{RWF-2000 Database and Proposed Method}
\subsection{Data Collection}
\label{data_collection}
To make violence detection more practical in realistic applications, we collect a new Real-World Fighting (RWF) dataset from the YouTube platform, which consists of 2,000 video clips captured by surveillance cameras in real-world scenes. 

Figure \ref{pipline} illustrates the pipeline of data collection. Firstly we search on the YouTube website by specifying a set of keywords related to violence (e.g., real fights, violence under surveillance, violent events), and obtain a list of URLs. We then employ a program to automatically download videos from the obtained links and check each video to remove irrelevant ones. To exhaustively extend the diversity of our dataset, we repeat the above procedures by changing keywords to various languages, including English, Chinese, German, French, Japanese, Russian, etc. Our selection criteria do not limit the kind of violence, including any form of subjectively identified violent activities (e.g., fighting, robbery, explosion, shooting, blood, assault). After these, we obtain many raw videos and cut each video into a 5-second clip with 30 FPS. In the end, we elaborately delete the noisy clips which contain unrealistic and non-monitoring scenes and annotate each clip as Violent or Non-Violent. 

Furthermore, as a total of 2,000 video clips are extracted from around 1,000 raw surveillance videos with extended footage, we take a set of means to avoid data leakage in splitting 'train/test' partitions. First, we employ a dictionary to record the ID of source surveillance video for each extracted clip. Second, a python script is used to allocate clips to the training set and test set randomly. Meanwhile, it guarantees that clips from the same source will not be assigned to the same collection by checking the dictionary records. Besides, we calculate the color histogram of each clip as a feature vector to compute the mutual cosine similarity between clips in the same collection. Based on the similarity ranking from high to low, we manually check the top 30\% of the data to ensure the reliability of the dataset partition. 

The proposed dataset has 2,000 video clips, split into two parts: the training set (80\%) and the test set (20\%). Half of the videos include violent behaviors, while others belong to non-violent activities. Table \ref{previous_datasets} thoroughly compares our RWF-2000 dataset with others. Our highlight is that surveillance cameras capture all videos in this dataset, and multimedia technologies modify none of them. The videos are close to real violent events. Besides, the number of videos also exceeds previous datasets. 

\begin{figure}[t]
  \centering
  \includegraphics[width=\linewidth]{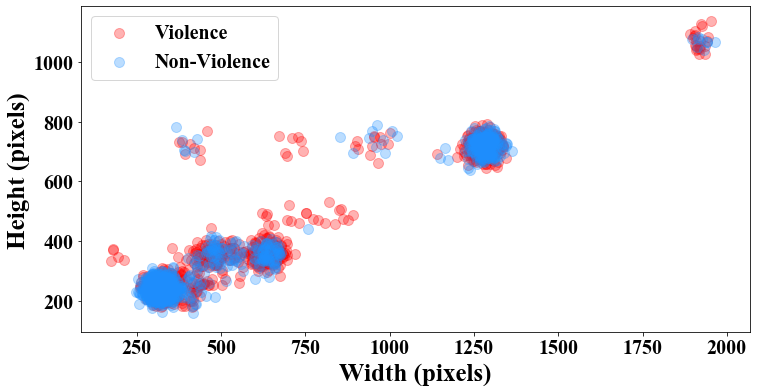}
  \caption{Resolution Distribution of the RWF-2000 Database.}
  \label{resolution}
\end{figure}

Figure \ref{resolution} demonstrates the distribution of video quality in the RWF-2000 dataset. As common camera settings, videos usually follow a  series of resolution standards (e.g., 720P, 1080P, 2K, 4K). Many data points in the distribution map will overlap. Therefore, we add a few random jitters to the original resolution data to obtain a better visualization. It is shown that several significant clusters are around the standard formats of 240P, 320P, 720P, and 1080P, respectively. Furthermore, the proportion of violent videos is almost uniform at different scales.

\subsection{Flow Gated Network}
\label{model_structure}
Most previous methods explore to extract appearance features from individual frames and then fuse them to model temporal information. Ng et al. \cite{r12} summarize various architectures for temporal feature pooling, and most of them are human-designed and tested one by one. As motion information may be useless due to the coarse pooling mechanism, we aim to design a temporal pooling mechanism achieved by network self-learning.

\begin{figure*}[t]
  \centering
  \includegraphics[width=\textwidth]{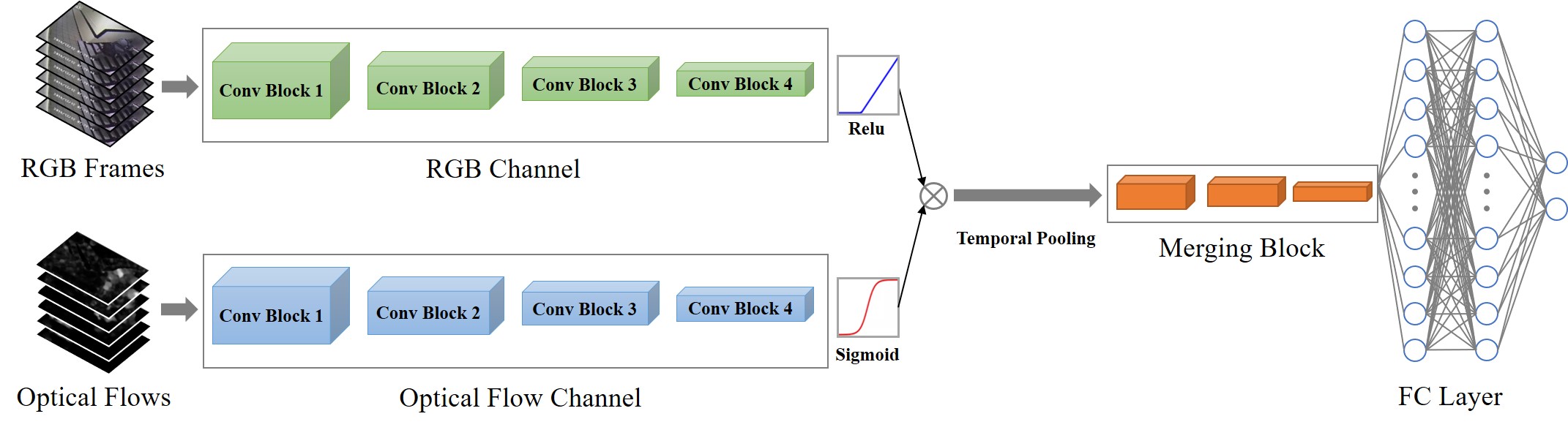}
  \caption{The structure of the Flow Gated Network.}
  \label{structure}
\end{figure*}

Figure \ref{structure} shows the structure of our proposed model with four parts: the RGB channel, the Optical Flow channel, the Merging Block, and the Fully Connected Layer. RGB channel and Optical Flow channel consist of cascaded 3D CNNs, and they have consistent structures so that their output could be fused. Merging Block is also composed of basic 3D CNNs, which process information after self-learned temporal pooling. Finally, the fully-connected layers generate output. Furthermore, we adopt the concept of depth-wise separable convolutions from MobileNet \cite{mobilenets} and Pseudo-3D Residual Networks \cite{p3d} to modify the 3D convolutional layers in our model, which can significantly reduce the model parameters without performance loss.

The highlight of this model is to utilize a branch of the optical flow channel to help build a pooling mechanism. Relu activation is adopted at the end of the RGB channel, while the sigmoid function is placed at the end of the Optical Flow channel. Then, outputs from RGB and Optical Flow channels are multiplied together and processed by a temporal max-pooling. Since the output of the sigmoid function is between 0 and 1, it is a scaling factor to adjust the output of the RGB channel. Meanwhile, as max-pooling only can reserve local maximum, the outcome of the RGB channel multiplied by one will have a larger probability to be retained, and the value multiplied by zero is more natural to be dropped. This mechanism is a kind of self-learned pooling strategy, which utilizes a branch of optical flow as a gate to determine what information the model should preserve or drop. The detailed parameters of the model structure are described in Table \ref{params}.

\section{Experiments}

\begin{figure}[t]
  \centering
  \includegraphics[width=\linewidth]{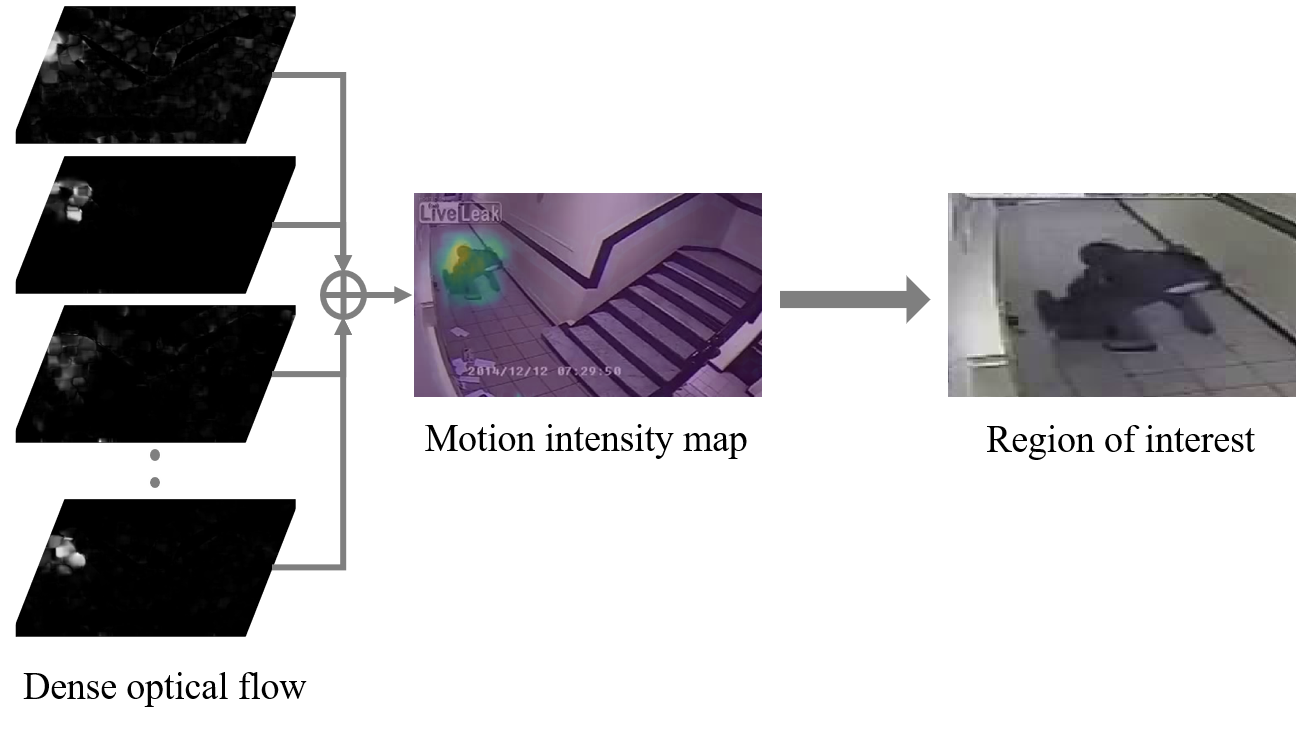}
  \caption{Cropping strategy using dense optical flow.}
  \label{Crop}
\end{figure}

\subsection{Basic Settings}
It is known that consecutive frames in a video are highly correlated, and the region of interest for recognizing human activity is usually a small area. We implement both cropping and sampling strategies to reduce the amount of input video data.

In the latency-intensive situation, there are methods to accelerate the computation of optical flow on different hardware devices (e.g., GPU \cite{gpu_based_opt_2}, FPGA \cite{gpu_based_opt_1}). While in this offline experimental case, we prefer a simple manner to complete this task by tools in OpenCV \cite{opencv}. Firstly, we employ Gunner Farneback's method \cite{optflow} to compute the dense optical flow between neighboring frames. The computed dense optical flow is a field of the 2-D displacement vector. Thus we calculate the norm of each vector to obtain a heat map for indicating the motion intensity. We use the sum of all the heat maps to be a final motion intensity map. The region of interest is extracted from the location with the most significant motion intensity (shown in Figure \ref{Crop}). Secondly, we sparsely sample frames from the video by a uniform interval for each input video and then generate a fixed-length video clip. By adopting both cropping and sampling, the amount of input data decreases significantly. In this project, the target length of video clips is 64, and the size of the cropped regions is 224 $\times$ 224. After sampling and cropping processes, the input data has a shape of 64 $\times$ 224 $\times$ 224 $\times$ 5. The last dimension has five channels containing three RGB channels and two optical flow channels (a horizontal component and a vertical component).

We implement the SGD optimizer with momentum (0.9) and the learning rate decay (1e-6). Also, we introduce the brightness transformation and random rotation as tricks for data augmentation, which help mitigate the over-fitting problem by simulating various lighting conditions and camera angles in real-world scenes.

\begin{table}[t]
  \centering
  \renewcommand\arraystretch{1.24}
  \caption{Parameters of the model architecture (the 'T' represents the number of repeats)}
  \label{params}
  \setlength{\tabcolsep}{4.5mm}{
  \begin{tabular}{|c|c|c|c|}
  \hline
    \textbf{Block Name} & \textbf{Type} &  \textbf{Filter Shape} & \textbf{T}\\ 
    \hline
             & Conv3d    & 1$\times$3$\times$3@16  &   \\
             & Conv3d    & 3$\times$1$\times$1@16  & 2 \\
    RGB/Flow & MaxPool3d & 1$\times$2$\times$2     &   \\
             \cline{2-4}
    Channels & Conv3d    & 1$\times$3$\times$3@32  &   \\
             & Conv3d    & 3$\times$1$\times$1@32  & 2 \\  
             & MaxPool3d & 1$\times$2$\times$2     &   \\
    \hline
    Fusion and & Multiply& None                    & 1 \\
    Pooling    & MaxPool3d& 8$\times$1$\times$1    & 1 \\
    \hline
             & Conv3d    & 1$\times$3$\times$3@64  &   \\
             & Conv3d    & 3$\times$1$\times$1@64  & 2 \\  
    Merging  & MaxPool3d & 2$\times$2$\times$2     &   \\         
             \cline{2-4}
    Block    & Conv3d    & 1$\times$3$\times$3@128 &   \\
             & Conv3d    & 3$\times$1$\times$1@128 & 1 \\  
             & MaxPool3d & 2$\times$2$\times$2     &  \\  
    \hline
    Fully-connected & FC layer  & 128           & 2 \\
            \cline{2-4}
     Layers     & Softmax   & 2             & 1   \\
    \hline
\end{tabular}}
\end{table}

\begin{table}[t]
  \centering
  \renewcommand\arraystretch{1.24}
  \caption{Evaluation of the proposed Flow Gated Network on the RWF-2000 dataset}
  \label{rwf_eval}
  \setlength{\tabcolsep}{2.0mm}{
  \begin{tabular}{|c|c|c|c|}
  \hline
    \textbf{Method} & \textbf{Train Accuracy(\%)}  & \textbf{Test Accuracy(\%)} & \textbf{Params} \\
    \hline
    RGB Only     & 89.50 & 84.50 & 248,402 \\
    OPT Only     & 82.31 & 75.50 & 248,258 \\
    Fusion (P3D) & 88.44 & 87.25 & 272,690 \\
    Fusion (C3D) & 96.50 & 85.75 & 507,154 \\
  \hline
\end{tabular}}
\end{table}

\subsection{Ablation Experiments}
In this part, we present detailed ablation studies to evaluate our model performance on the proposed RWF-2000 dataset. Table \ref{rwf_eval} shows the experimental results of four different cases. 

The \emph{RGB Only} represents removing the optical flow channel while only keeping the RGB channel as input. On the other hand, the \emph{OPT Only} shows removing the RGB channel and only keeping the optical flow channel. Subsequently, the \emph{Fusion (P3D)} proves that the fused version of our proposed method can achieve the best accuracy of 87.25\%.  

To compare the tradeoff of computation amount between depth-wise separable 3D convolutions and traditional 3D convolutions, we also test the fused version of our proposed method without adopting the concept from P3D-Net \cite{p3d}. The \emph{Fusion (C3D)} shows the case of utilizing traditional 3D convolutions, instead of depth-wise separable 3D convolutions. This model achieves an accuracy of 85.75\%, with nearly double the amount of model parameters. This result shows that modifying the standard 3D convolutions to the depth-wise separable 3D convolutions could greatly reduce the model parameters with increasing the classification performance.

\subsection{Comparisons}
Although we have listed a series of datasets for violence detection in Section \ref{other_datasets}, it is still difficult to evaluate a method on all of them due to their heterogeneity in data format, annotation type, etc. Therefore, we only test our proposed method on the Hockey Fight, the Movies Fight, and the Crowd Violence datasets, all consistent with the RWF-2000 dataset in data format, annotation type, and close video footage. Table \ref{other_eval} shows comparisons between the Flow Gated Network and other methods. The authors report performances of the first eight methods (from ViF to 3D ConvNet) in their papers. The last six methods are implemented and trained by ourselves due to the lack of existing experiments on these datasets. Generally, deep learning based approaches significantly outperform methods which depend on hand-crafted features. 

\begin{figure}[t]
\centering
\label{opt_case}
\subfigure[Example 1]{
\begin{minipage}[t]{0.48\linewidth}
\centering
\includegraphics[width=1.5in]{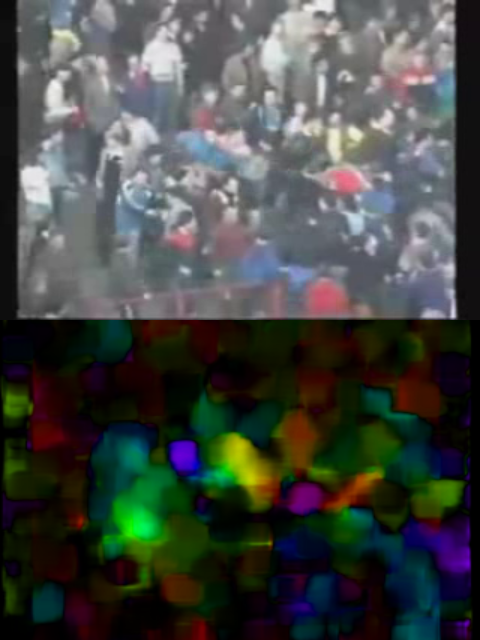}
\end{minipage}%
}%
\subfigure[Example 2]{
\begin{minipage}[t]{0.48\linewidth}
\centering
\includegraphics[width=1.5in]{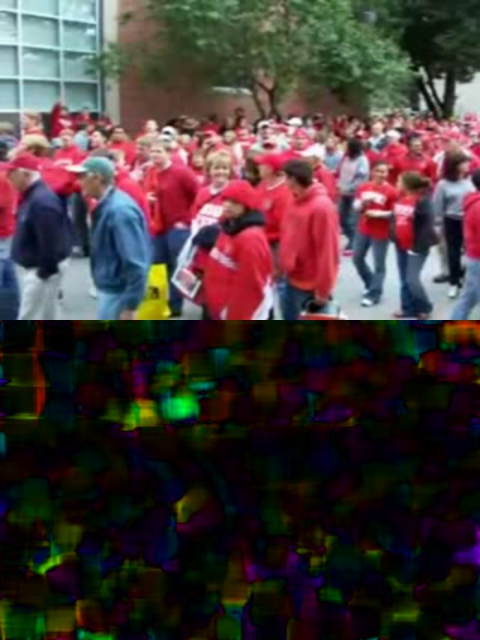}
\end{minipage}%
}%
\caption{Bad cases of computing optical flow in the Crowd Violence dataset.}
\end{figure}

Table \ref{other_eval} reveals two experimental results. First, the deep learning based methods are all over-fitted on the Movies Fight dataset. Hence, this dataset is no longer suitable to be used in the current situation. Second, The performance of most methods carrying optical flow information has a significant drop in the Crowd Violence dataset. The reason for this phenomenon is due to the computing error of optical flow. There are three premise assumptions \cite{opt_computation} of computing optical flow: the brightness between adjacent frames is constant; the time between the adjacent frames is continuous, or the motion change is small; pixels in the same sub-image have the same motion.  While in terms of camera movements, stable lighting conditions, and low image quality, videos in the Crowd Violence dataset are far from the requirements. Figure \ref{opt_case} shows two bad cases of computing the optical flow in the Crowd Violence dataset, which can visualize the reason why flow-based methods perform poorly in this situation.

\begin{table}[!t]
  \centering
  \renewcommand\arraystretch{1.25}
  \caption{Comparisons between the proposed method and others on the previous datasets}
  \label{other_eval}
  \setlength{\tabcolsep}{2.0mm}{
  \begin{tabular}{|c|c|c|c|c|}
    \hline
    \textbf{Type} &\textbf{Method} & \textbf{Movies}  & \textbf{Hockey} & \textbf{Crowd} \\
   
     \hline
     \multirowcell{6}{Hand-Crafted\\Features}
      & ViF \cite{r3}                     &  -      & 82.90\% & 81.30\%  \\ 
      &LHOG+LOF \cite{zhou2018violence}   &  -      & 95.10\% & 94.31\%  \\ 
      &HOF+HIK \cite{r9}                  & 59.0\%  & 88.60\% & -        \\
      &HOG+HIK \cite{r9}                  & 49.0\%  & 91.70\% & -        \\
      &MoWLD+BoW    \cite{MoWLD}          &  -      & 91.90\% & 82.56\%  \\
      &MoSIFT+HIK    \cite{r9}            & 89.5\%  & 90.90\% & -        \\
      \cline{1-1}
      \multirowcell{6}{Deep-Learning\\Based}
      &FightNet \cite{zhou2017violent}    & 100\%   & 97.00\% &  -       \\
      &3D ConvNet \cite{song2019novel}    & 99.97\% & 99.62\% & 94.30\%  \\ 
      &ConvLSTM \cite{convlstm-violence}  & 100\%   & 97.10\% & 94.57.   \\
      &C3D \cite{r1}                      & 100\%   & 96.50\% & 84.44\%  \\
      &I3D(RGB only) \cite{i3d}           & 100\%   & 98.50\% & 86.67\%  \\
      &I3D(Flow only) \cite{i3d}          & 100\%   & 84.00\% & 88.89\%  \\
      &I3D(Fusion) \cite{i3d}             & 100\%   & 97.50\% & 88.89\%  \\  
      &Ours                               & 100\%   & 98.00\% & 88.87\%  \\
    \hline
\end{tabular}}
\end{table}

\begin{table}[!t]
  \centering
  \renewcommand\arraystretch{1.25}
  \caption{Comparisons between the proposed method and others on the RWF-2000 dataset}
  \label{other_eval_1}
  \setlength{\tabcolsep}{5.1mm}{
  \begin{tabular}{|l|c|c|}
  \hline
    \textbf{Method} & \textbf{Accuracy(\%)}  & \textbf{Params (M)} \\
    \hline
      ConvLSTM \cite{convlstm-violence} & 77.00 & 47.4 \\
      C3D \cite{r1}                     & 82.75 & 94.8 \\
      I3D (RGB only) \cite{i3d}          & 85.75 & 12.3 \\
      I3D (Flow only) \cite{i3d}         & 75.50 & 12.3 \\
      I3D (TwoStream) \cite{i3d}         & 81.50 & 24.6 \\  
      Ours (best version)               & 87.25 & 0.27 \\
  \hline
\end{tabular}}
\end{table}

Compared to the generic human action recognition, the essence of video-based violence detection is a data-driven problem. Hence, we compare our method with others presented for violence detection and test some classical models for generic action recognition on different violence datasets. For a fair comparison, we train our model and others without any pre-training procedures.  Table \ref{other_eval_1} shows benchmarks of the RWF-2000 dataset. Our proposed method achieves an accurate result for violence recognition. Meanwhile, it has fewer trainable model parameters.

\section{Conclusion}
This paper presents both a novel dataset and a method for violence detection in surveillance videos. The proposed RWF-2000 dataset is the currently largest surveillance video dataset used for violence detection in realistic scenes. Moreover, a unique pooling mechanism is employed by optical flow, which could implement temporal feature pooling instead of human-designed strategies. In the future, we will explore methods without utilizing optical flow explicitly. Also, we will proceed to expand the size of the RWF-2000 dataset.

\section*{Acknowledgment}
This research is funded in part by the National Natural Science Foundation of China (61773413), Key Research and Development Program of Jiangsu Province (BE2019054), Six talent peaks project in Jiangsu Province (JY-074), Guangzhou Municipal People's Livelihood Science and Technology Plan (201903010040).

\bibliographystyle{IEEEtran}
\bibliography{references}

\end{document}